\pdfoutput=1
\documentclass[11pt]{article}
\usepackage{EMNLP2022}
\usepackage{times}
\usepackage{latexsym}
\usepackage{graphicx}
\usepackage{times}
\usepackage{latexsym}
\usepackage{mathtools}
\usepackage{amsmath}
\usepackage{amsfonts}
\usepackage{multirow}
\usepackage{booktabs}
\usepackage{colortbl}
\usepackage{longtable}
\usepackage{hyperref} 
\usepackage{float}
\usepackage{subcaption}
\usepackage[T1]{fontenc}
\usepackage[utf8]{inputenc}

\usepackage{microtype}

\usepackage{inconsolata}

\def\Snospace~{\S{}} % If i don't add this overleaf complains

\title{How sensitive are translation systems to extra contexts? Mitigating gender bias in Neural Machine Translation models through relevant contexts}

\author{Shanya Sharma \\
  Walmart Labs, India \\
  \texttt{shanya.sharma@walmart.com} \\\And
  Manan Dey \\
  SAP Labs, India \\
  \texttt{manan.dey@sap.com} \\ \AND
  Koustuv Sinha \\
  McGill University, Montreal, Canada, \\
  Mila - Quebec AI Institute, \\
  \texttt{koustuv.sinha@mail.mcgill.ca} \\
  }

\begin{document}
\maketitle
\begin{abstract}
Neural Machine Translation systems built on top of Transformer-based architectures are routinely improving the state-of-the-art in translation quality according to word-overlap metrics. However, a growing number of studies also highlight the inherent gender bias that these models incorporate during training, which reflects poorly in their translations. In this work, we investigate whether these models can be instructed to fix their bias during inference using targeted, guided instructions as contexts. By translating relevant contextual sentences \textit{during inference} along with the input, we observe large improvements in reducing the gender bias in translations, across three popular test suites (WinoMT, BUG, SimpleGen). We further propose a novel metric to assess several large pre-trained models (OPUS-MT, M2M-100) on their sensitivity towards using contexts during translation to correct their biases. Our approach requires no fine-tuning and thus can be used easily in production systems to de-bias translations from stereotypical gender-occupation bias \footnote{Our evaluation data and code are publicly available at \href{https://github.com/manandey/bias_machine_translation}{https://github.com/manandey/bias\_machine\_translation}}. We hope our method, along with our metric, can be used to build better, bias-free translation systems.
\end{abstract}

\section{Introduction}

\begin{figure}[t]
  \includegraphics[trim={200 100 150 120},clip,width=0.5\textwidth]{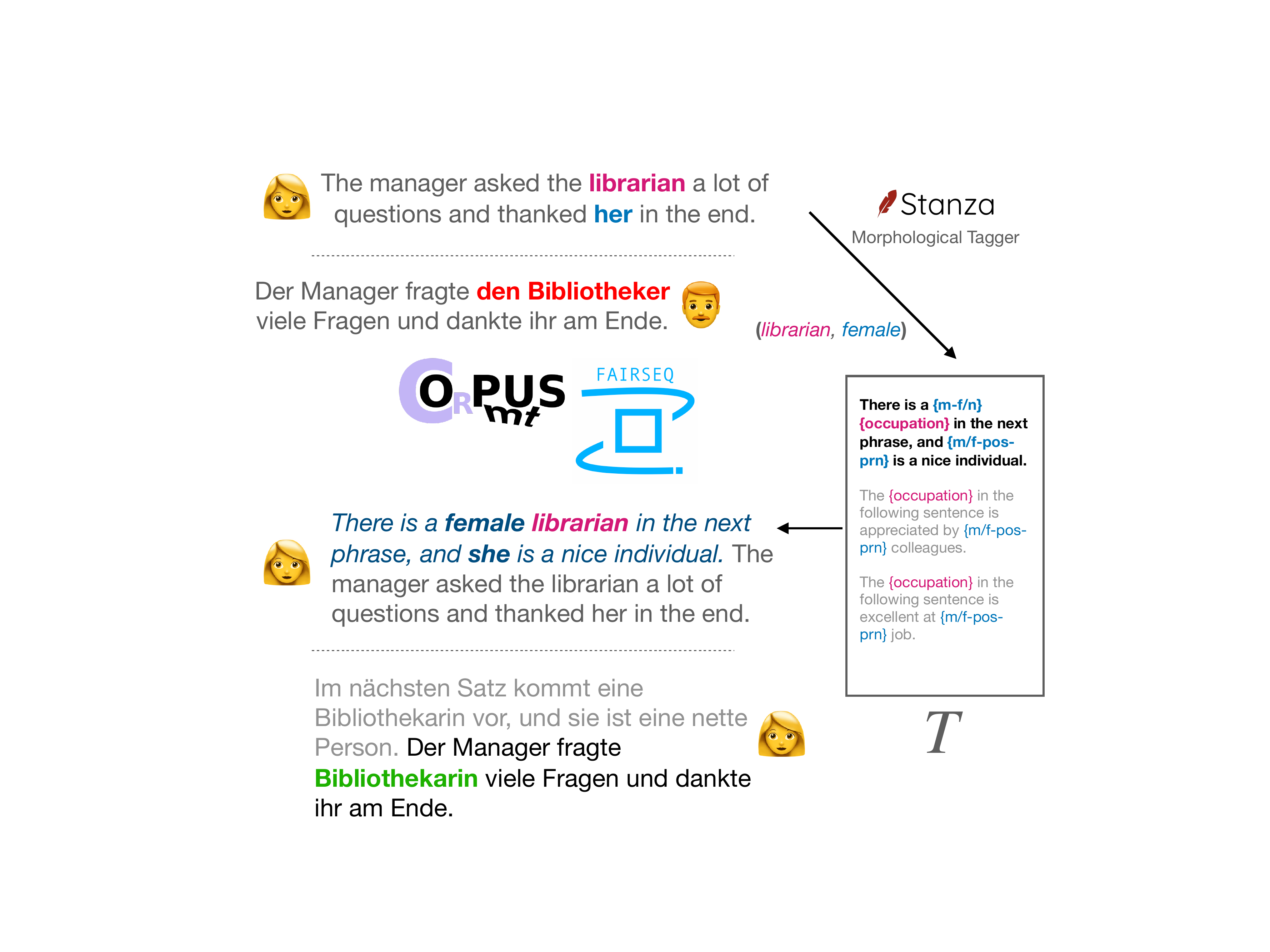}
  \caption{An example of our de-biasing pipeline, using OPUS-MT \cite{tiedemann-thottingal-2020-opus} model on English to German translation on a sample drawn from WinoMT \cite{stanovsky-etal-2019-evaluating} dataset. We correct the bias during inference by providing additional relevant \textit{context} (built using our template bank $T$) to the input, which we remove after the translation.}
  \label{fig:example}
\end{figure}

Despite the ongoing success of large pre-trained Transformer \cite{vaswani2017} based models in Neural Machine Translation (NMT), these systems are immensely prone to various forms of gender biases in their learned representations. 
Recent work \cite{stanovsky-etal-2019-evaluating,prates2020assessing}  has found out that a specific kind of gender bias exists in the translation of NMT models. Specifically, sentences containing stereotypical occupations\footnote{Occupations obtained from the US Bureau of Labor Statistics, \href{http://bls.gov/cps/cpsaat11.htm}{http://bls.gov/cps/cpsaat11.htm}} which are typically gender-unbalanced in the training data are translated per their respective stereotypes (e.g., \textit{nurse} tends to be associated to \textit{female} pronouns) intact in the output (\autoref{fig:example}).
It is, therefore, imperative to study effective de-biasing techniques to instruct a translation model to output unbiased translations.

De-biasing biased gender associations have been thoroughly investigated in the light of static word embeddings \cite{bolukbasi,zhao2018a,elazar-goldberg-2018-adversarial}. Relatively fewer works exist in de-biasing on contextual word embeddings (such as Transformer-based models). The prevalent approach is to fine-tune a pre-trained contextualized embedding while balancing gender-specific associations \cite{zhao2019}, or by fine-tuning the word embeddings of the pre-trained model itself \cite{kaneko-bollegala-2021-debiasing}. However, in the case of NMT, fine-tuning is expensive as it requires massive parallel corpora, and to this date, we do not have a massive, gender-balanced parallel corpora to begin with.

Thus, in this work, we investigate whether context can be leveraged as a way to improve the bias of a translation system. 
NMT systems have been reported to be highly sensitive to input sentences \cite{fadaee-monz-2020-unreasonable, dankers-etal-2022-paradox}. In this work, we aim to use it to our advantage to de-bias a model. 
Instead of fine-tuning a translation model, we embark on improving the generation by allowing the model to focus on contextual information during inference. 
Concretely, we expose the model to unambiguous context alongside to the input to translate, containing the unbiased gender association of the entity in the input\footnote{Unlike \citet{fadaee-monz-2020-unreasonable}, we do not modify the source sentence in itself, instead we append or prepend extra contextual information, which can also be compared to tuning-free prompt mining setup in the literature \cite{prompt_survey21}.}. Specifically, we do not modify the input: instead we either \textit{prepend} or \textit{append} the context to the input as a separate sentence, separated by delimiters (\autoref{fig:example}). We observe that using this context, state-of-the-art translation models are able to reduce their gender association biases for occupations considerably in the output translation, purely during inference. This improvement varies according to model, language-pair and occupations.

Concretely, in this work we systematically study the effect of contexts in the source language (English) affecting the  occupation gender-bias in the target translation (German, French, and Spanish) using two popular state-of-the-art NMT models: M2M-100 \cite{JMLR:v22:20-1307} and OPUS-MT \cite{tiedemann-thottingal-2020-opus}, in three publicly available occupation-gender bias datasets: WinoMT \cite{stanovsky-etal-2019-evaluating}, BUG \cite{levy-etal-2021-collecting-large} and SimpleGen \cite{renduchintala-williams-2022-investigating}.
% We further measure the translation models' sensitivity to additional relevant contexts using a novel Context Sensitivity Score.
We find that both models can be de-biased significantly by adding unambiguous, relevant contexts, with the largest improvement being that of M2M-100, which exhibits higher sensitivity towards additional contexts. Our proposed method thus introduces a simple and effective way to de-bias translations during inference.

\section{Approach}
\label{sec:approach}

To mitigate gender-stereotyped bias for occupations in NMT, we consider the approach of adding a context to the input, as a separate sentence either prepended or appended with the original input sentence. This context is generated using handcrafted templates, which contain unambiguous signals about the gender of the profession in the source sentence. In particular, we investigate the capability of NMT models to extract signals from these contexts and mitigate gender bias in the translation during inference. Our approach can also be thought of as an in-context learning scheme, as popularized by \citet{Brown2020:GPT3}.

\subsection{Template construction and usage}
\label{sec:template_construction}

We start with a parallel corpora $D$ consisting of sentence pairs $(X,Y)$, where each source sentence $X$ contains a target entity of gender \textit{male} or \textit{female}\footnote{We acknowledge that a limitation of this work is that we do not consider the non-binary genders, and we leave it for potential future work to explore the context sensitivity of non-binary genders in language. We also hope our work will lead to new ideas and better methods for mitigating biases about non-binary and transgender people in the future.} and associated with an occupation. $X$ also contain gender-specific pronoun(s) to indicate the gender of the target entity. $Y$ denotes the gold translation of $X$.
We use a pre-trained translation model to translate $X$ to $\hat{Y}$. This translation permeates the stereotypical bias of the target entity in the source, which we aim to fix in this work, by providing a context during translation.

To construct unambiguous contexts, we carefully create templates $t$, which can be used to generate a contextual sentence, $c$. This context provides enough signal unambiguously to convey the correct gender of the target entity in $X$ (\autoref{fig:example}). For example, given a context template $t = $ \textit{``The }\{occupation\} \textit{in the next sentence identifies }\{male or female self-reference pronoun\} \textit{using the pronouns }\{male or female subject pronoun\}/\{male or female object pronoun\}'', and given an occupation gender pair \texttt{(nurse, male)}, we construct the following context $c$: \textit{``The nurse in the next sentence identifies himself using the pronouns he/him.''}

Thus, given the input sentence $X$, we prepend or append the context $c$ to construct a new input for translation, $X_c = [c \| X]$ or $X_c = [X \| c]$, where $\|$ is the delimiter which separates the two sentences. We translate this sentence $X_c$ to a target language $\mathcal{L}$ using the pre-trained translation model to output the translation $\hat{Y}_c$. To extract the intended translation, we drop the translated context by splitting the output using the delimiter $\|$, to get $\hat{Y}_{\not c}$, which is the translation of $X_c$ after removing $c$.

\subsection{Choosing a template}
\label{sec:greedy_strategy}

Using our template construction strategy, we create $T$ unique templates which could be applied to a given input sentence.
We use a greedy strategy to choose a template to apply for a given sentence $X$.
Following the formulation of \citet{stanovsky-etal-2019-evaluating}, we first use a heuristic morphological tagger to extract the gender of the target entity from the source ($g_X$) and from the translation ($g_{\hat{Y}}$). We use Stanza \cite{qi-etal-2020-stanza} as the morphological tagger and \texttt{AWeSOME} aligner \cite{dou-neubig-2021-word} to align the source and target entities. $g_X \ne g_{\hat{Y}}$ indicates the presence of stereotypical bias \footnote{Note, we do not use the ground truth annotation $g_Y$ to decide this, as it is unavailable during testing. Our method relies on the accuracy of the heuristic morphological tagger.}. In those sentences, we iteratively search for a relevant context $c, \forall t \in T$ such that $g_X=g_{\hat{Y}_{\not c}}$. We stop this search once we exhaust our set of templates in $T$.

\subsection{Experiment Details}

\noindent\textbf{Models}: For our experiments, we consider using the two most commonly used open-source multilingual translation models, M2M-100 and OPUS-MT, for evaluation. M2M-100 \citep{JMLR:v22:20-1307} which is a many-to-many multilingual encoder-decoder translation model that can translate directly between any pair of 100 languages, based on the Transformer\cite{vaswani2017} architecture.
We use the 418 Million parameters version of the model.
OPUS-MT  \citep{tiedemann-thottingal-2020-opus} is a collection of bilingual and multilingual models based on the standard 6-layer 8-head Transformer architecture. We use HuggingFace \cite{huggingface} model hub to load and run inference for both of the above models.

\noindent\textbf{Languages \& Datasets}: We perform our experiments using translations from English to three target languages: German, French and Spanish. We chose these three languages as they are well-supported by Stanza for performing morphological analysis, while also being supported by the NMT models we consider above. For each of these languages, we carry the evaluation on three datasets, WinoMT \cite{stanovsky-etal-2019-evaluating}, BUG \citep{levy-etal-2021-collecting-large} and SimpleGen \citep{renduchintala-williams-2022-investigating}.

\noindent\textbf{Generating templates:} We construct $|T|=87$ unique templates\footnote{\autoref{tab:contexts-table} in the Appendix contains a full list of templates.} with varying linguistic properties.
% We generate diverse set of templates based on the procedure described in \autoref{sec:template_construction}.
For each model and language-pair, we prune a subset of templates that evoke stereotypical gender biases in their translations.
%We prune this set of templates per model and language-pair to remove the ones exhibiting stereotypical occupation-gender biases in their own translations. 
We then apply these templates to the input sentences in the dataset and translate the combined sentence.
We observe that the choice of delimiter used to combine the input sentence with the context has a significant effect on the actual translation quality of the input sentence. In our primary experiments, we choose hash(\#) as our delimiter since it provides a substantial improvement in the bias while also ensuring minimal change in translation quality. We discuss more about the choice and impact of delimiters in \autoref{sec:abl_bleu}.

\noindent\textbf{Evaluation.} To compute the gender translation \textit{accuracy}, we extract the predicted gender from the translation, $g_{\hat{Y}}$, using the morphological tagger. Then, we measure if this predicted gender $g_{\hat{Y}}$ is the same as that of an annotated, gold truth gender $g_{Y}$ for the same entity. We use BLEU scores to evaluate the translation of the combined sentence $X_c$ using the same setup, which contains the source sentence $X$ and a context $c$. In this case, post translation, we drop the translated output of $c$ and evaluate $Y_{\not c}$ using the same method as described above.

\section{Results \& Analysis}

In this section, we conduct a series of experiments and analysis to understand the viability of correcting gender bias in translation by using contexts.

\begin{table*}
	\centering
	\resizebox{\linewidth}{!}{%
		\begin{tabular}{c|c|c|c|ccccc} 
			\hline
			&                           &                                   & Without Context                      & \multicolumn{5}{c}{With   Context}                                                                                                     \\\cline{4-9}   
			\multirow{-2}{*}{Dataset}   & \multirow{-2}{*}{Model}   & \multirow{-2}{*}{Target Language} & $\mathcal{A} (\%)$                    & $\mathcal{A}_C (\%)$                               & $\mathcal{A}_{all}~(\%) $                      & $\mathcal{CSS}$                  & \multicolumn{1}{l}{$\mathcal{C}_U$} & \multicolumn{1}{l}{$\mathcal{C}_L$} \\ \hline
			                            &                           & German (de)                       & 60.57 (8.64)                          & \multicolumn{1}{c} {\textbf{\color{teal}{82.13}}} & \multicolumn{1}{c}{\textbf{62.09 (9.81)}}   & \multicolumn{1}{c}{0.27 (0.38)} & \multicolumn{1}{c}{54.60}           & 10.64                                \\  
			&                           & French (fr)                 & 57.70 (12.20)                         
			& \multicolumn{1}{c}{\textbf{\color{teal}{73.64}}}  & \multicolumn{1}{c}{\textbf{59.88 (13.67)}} & \multicolumn{1}{c}{0.34 (0.43)}  & \multicolumn{1}{c}{37.60}  & 3.80   \\  
			                            & \multirow{-3}{*}{OPUS-MT} & Spanish (es)                      & 60.10 (10.18)                         & \multicolumn{1}{c}{\textbf{\color{teal}{76.79}}}  & \multicolumn{1}{c}{58.97 (8.64)}            & \multicolumn{1}{c}{0.31 (0.41)} & \multicolumn{1}{c}{41.80}           & 9.96                                 \\ \cline{2-9} 
			                            &                           & German (de)                       & 58.71 (10.82)                         & \multicolumn{1}{c}{\textbf{\color{teal}{80.65}}}  & \multicolumn{1}{c}{58.25 (10.86)}           & \multicolumn{1}{c}{0.25 (0.35)} & \multicolumn{1}{c}{53.13}           & 6.30                                 \\  
			                            &                           & French (fr)                       & 49.43 (19.09)                         & \multicolumn{1}{c}{\textbf{\color{teal}{76.83}}}  & \multicolumn{1}{c}{\textbf{54.03 (7.42)}}   & \multicolumn{1}{c}{0.37 (0.44)} & \multicolumn{1}{c}{54.18}           & 17.40                                \\  
			\multirow{-6}{*}{WinoMT}    & \multirow{-3}{*}{M2M-100} & Spanish (es)                      & 56.56 (10.23)                         & \multicolumn{1}{c}{\textbf{\color{teal}{85.35}}}  & \multicolumn{1}{c}{\textbf{59.25 (11.69)}}  & \multicolumn{1}{c}{0.35 (0.41)} & \multicolumn{1}{c}{66.27}           & 18.45                                \\ \hline
			                            &                           & German (de)                       & 70.72 (17.08)                         & \multicolumn{1}{c}{\textbf{\color{teal}{85.60}}}  & \multicolumn{1}{c}{66.91 (16.09)}           & \multicolumn{1}{c}{0.20 (0.30)} & \multicolumn{1}{c}{51.43}           & 8.40                                 \\  
			                            &                           & French (fr)                       & 55.96 (19.27)                         & \multicolumn{1}{c}{\textbf{\color{teal}{75.90}}}  & \multicolumn{1}{c}{\textbf{61.34 (16.91)}}  & \multicolumn{1}{c}{0.22 (0.35)} & \multicolumn{1}{c}{45.70}           & 11.00                                \\  
			                            & \multirow{-3}{*}{OPUS-MT} & Spanish (es)                      & 74.85 (21.64)                         & \multicolumn{1}{c}{\textbf{\color{teal}{86.74}}}  & \multicolumn{1}{c}{\textbf{75.50  (18.31)}} & \multicolumn{1}{c}{0.16 (0.29)} & \multicolumn{1}{c}{47.72}           & 7.20                                 \\ \cline{2-9} 
			                            &                           & German (de)                       & 58.13 (10.97)                         & \multicolumn{1}{c}{\textbf{\color{teal}{87.36}}}  & \multicolumn{1}{c}{\textbf{67.09 (15.20)}}  & \multicolumn{1}{c}{0.39 (0.42)} & \multicolumn{1}{c}{70.39}           & 25.95                                \\  
			                            &                           & French (fr)                       & 48.08 (16.89)                         & \multicolumn{1}{c}{\textbf{\color{teal}{78.84}}}  & \multicolumn{1}{c}{\textbf{57.92 (20.22)}}  & \multicolumn{1}{c}{0.40 (0.43)} & \multicolumn{1}{c}{59.76}           & 22.27                                \\  
			\multirow{-6}{*}{BUG}       & \multirow{-3}{*}{M2M-100} & Spanish (es)                      & 63.19(17.10)                          & \multicolumn{1}{c}{\textbf{\color{teal}{82.25}}}  & \multicolumn{1}{c}{\textbf{72.34(19.75)}}   & \multicolumn{1}{c}{0.32(0.40)}  & \multicolumn{1}{c}{61.11}           & 18.18                                \\ \hline
			                            &                           & German (de)                       & \cellcolor[HTML]{FFFFFF}58.03 (15.52) & \multicolumn{1}{c}{\textbf{\color{teal}{83.37}}}  & \multicolumn{1}{c}{\textbf{59.95 (15.57)}}  & \multicolumn{1}{c}{0.33 (0.40)} & \multicolumn{1}{c}{60.37}           & 6.70                                 \\  
			                            &                           & French (fr)                       & 57.28 (12.04)                         & \multicolumn{1}{c}{\textbf{\color{teal}{83.29}}}  & \multicolumn{1}{c}{\textbf{62.70 (12.37)}}  & \multicolumn{1}{c}{0.29 (0.39)} & \multicolumn{1}{c}{60.89}           & 16.90                                \\  
			                            & \multirow{-3}{*}{OPUS-MT} & Spanish (es)                      & 67.34 (9.70)                          & \multicolumn{1}{c}{\textbf{\color{teal}{86.48}}}  & \multicolumn{1}{c}{\textbf{70.65 (11.41)}}  & \multicolumn{1}{c}{0.21 (0.34)} & \multicolumn{1}{c}{58.62}           & 8.30                                 \\ \cline{2-9} 
			                            &                           & German (de)                       & 54.05 (5.72)                          & \multicolumn{1}{c}{\textbf{\color{teal}{79.84}}}  & \multicolumn{1}{c}{\textbf{56.10 (8.13)}}   & \multicolumn{1}{c}{0.29 (0.39)} & \multicolumn{1}{c}{56.12}           & 10.00                                \\  
			                            &                           & French (fr)                       & 53.34 (9.80)                          & \multicolumn{1}{c}{\textbf{\color{teal}{81.45}}}  & \multicolumn{1}{c}{\textbf{60.00 (8.45)}}   & \multicolumn{1}{c}{0.33 (0.41)} & \multicolumn{1}{c}{60.25}           & 13.75                                \\  
			\multirow{-6}{*}{SimpleGen} & \multirow{-3}{*}{M2M-100} & Spanish (es)                      & 59.75 (7.76)                          & \multicolumn{1}{c}{\textbf{\color{teal}{87.42}}}  & \multicolumn{1}{c}{\textbf{64.22 (11.49)}}  & \multicolumn{1}{c}{0.23 (0.32)} & \multicolumn{1}{c}{68.75}           & 5.20                                 \\  \hline
		\end{tabular}
	}
	\caption{Full results containing per dataset, per model and per language pairs. In ``Without Context", $\mathcal{A}$ reflects the accuracy of correct gender associations in the translation. In ``With Context", $\mathcal{A}_C$ is the accuracy of the overall dataset using the greedy approach. $\mathcal{A}_{all}$ reflects the average accuracy of correct gender associations for all templates applied on all sentences. The values in 'green' represent the highest accuracy score obtained by a language-model pair and the values in bold represent the values where average accuracy improved after applying all templates to the sentences. $\mathcal{CSS}$ represents the \texttt{CSS Score} (\autoref{sec:css}). The values in bracket represent the standard deviation for the corresponding metric. $\mathcal{C}_U$ and $\mathcal{C}_L$ represents respectively the percentage of the biased sentences where at least one template / all templates yields the correct prediction of the gender association. }
	\label{tab:full_table}
\end{table*}

\subsection{Does addition of templates allow the model to correct its bias?}

\noindent\textbf{Setup.} Concretely, we want the model to generate the appropriate gender-specific pronouns for the occupations in the translation $Y$ according to the gender $g_X$ in the input sentence $X$. We first apply our greedy template selection strategy to the WinoMT, BUG and SimpleGen datasets, using OPUS-MT and M2M-100 models. We compute the greedy search accuracy $\mathcal{A}_C$.
% We also investigate the fraction of the dataset where at least one context is able to rectify the bias, using the lower bound accuracy $\mathcal{C}_{L}$.

\noindent\textbf{Results.} We find that by the application of greedy strategy (\autoref{sec:greedy_strategy}), the accuracy $\mathcal{A}_C$ is significantly high (\autoref{tab:full_table}), with the highest performance improvement in BUG dataset (\textbf{87.36\%} for M2M-100 (German) compared with baseline 58.13\%). This is a promising result, as even accounting for morphological and heuristic gender detection approximation, it is possible to effectively de-bias a gender stereotype of a profession in translation by adding extra context.

% Following the results of $\mathcal{A}_C$ and $\mathcal{A}_{\text{all}}$, we further investigate the subset of the data where adding a template is useful, by constructing an upper and lower bound. Specifically, $\mathcal{C}_{U}$ denotes the fraction of the biased subset where at least one context is able to rectify the bias, subject to application of a context, and $\mathcal{C}_{L}$ denotes the fraction where \textit{all} templates $t \in T$ yield an unbiased translation.
% The highest lower bound $\mathcal{C}_L$ is also observed for the same dataset and language-model pair (25.95\%).

\noindent\textbf{Takeaway.} \textit{For most sentences, there exists at least one template which is able to correct the bias.}

\subsection{Is the non-greedy strategy also an effective method to reduce translation bias?}
\label{sec:average_acc}

\noindent\textbf{Setup.} Since the greedy strategy stops the search once it finds a working template, we also investigate a non-greedy strategy. Specifically, we naively apply all $T$ templates to all data points in $D$, and compute an average accuracy over $D \times T$. We denote this as the \textit{average accuracy}, $\mathcal{A}_{\text{all}}$.

\noindent\textbf{Results.} We also observe a marked improvement over averaged accuracy ($\mathcal{A}_{\text{all}}$) across most language-model pairs (\autoref{tab:full_table}) when we apply all templates from $T$, \textit{without} using our greedy template selection strategy. While we see slight performance dips in WinoMT, the least improvement is for BUG dataset using OPUS-MT model for German, where we see a significant decrease in performance after adding contexts. However, with M2M model, we observe the highest improvement in the same dataset across all languages. This result is possibly due to OPUS-MT German model being significantly worse in raw translation quality, as we observe in \autoref{tab:bleu-table}.

Unsurprisingly, we observe consistent improvement in SimpleGen dataset across all language-model pairs. Since the SimpleGen dataset is constructed from artificially generated data, it is effectively the least ambiguous among the three datasets \cite{renduchintala-williams-2022-investigating}, hence enabling the models to fully exploit the additional context.

\noindent\textbf{Takeaway.} \textit{On average, contexts correct the bias of translations across different datasets, language pairs and models.}

\subsection{Does the context impact the translation quality?}
\label{sec:abl_bleu}

\begin{table}
\resizebox{0.48\textwidth}{!}{%
\begin{tabular}{ccccc}
\hline
Target Language             & \multicolumn{2}{c}{OPUS-MT}          & \multicolumn{2}{c}{M2M-100}          \\ \hline
\multirow{4}{*}{German(de)} & \multicolumn{1}{c}{original}  & 45 & \multicolumn{1}{c}{original}  & 25 \\ 
                            & \multicolumn{1}{c}{hash (\#)}  & \textbf{45} & \multicolumn{1}{c}{hash (\#)}  & \textbf{25} \\
                            
                            & \multicolumn{1}{c}{period (.)}  & 44 & \multicolumn{1}{c}{period (.)}  & 23 \\
                            
                            & \multicolumn{1}{c}{colon (:)}     & 42 & \multicolumn{1}{c}{colon (:)}     & 24 \\
                            
                            & \multicolumn{1}{c}{semicolon (;) } & 44 & \multicolumn{1}{c}{semicolon (;) } & 24 \\ \hline
\multirow{4}{*}{French(fr)} & \multicolumn{1}{c}{original}  & 56 & \multicolumn{1}{c}{original}  & 37 \\ 
                            & \multicolumn{1}{c}{hash (\#)}  & \textbf{54} & \multicolumn{1}{c}{hash (\#)}  & \textbf{38} \\
                            
                            & \multicolumn{1}{c}{period (.)}  & 52 & \multicolumn{1}{c}{period (.)}  & 32 \\
                            
                            & \multicolumn{1}{c}{colon (:)}     & 53 & \multicolumn{1}{c}{colon (:)}     & 35 \\ 
                            
                            & \multicolumn{1}{c}{semicolon (;) } & 53 & \multicolumn{1}{c}{semicolon (;) } & 35 \\ \hline
\multirow{4}{*}{Spanish(es)} & \multicolumn{1}{c}{original} & 62 & \multicolumn{1}{c}{original} & 42 \\ 
                            & \multicolumn{1}{c}{hash (\#)}  & \textbf{62} & \multicolumn{1}{c}{hash (\#)}  & \textbf{42} \\
                            
                            & \multicolumn{1}{c}{period (.)}  & \color{red}{53} & \multicolumn{1}{c}{period (.)}  & \color{red}{34} \\
                            
                            & \multicolumn{1}{c}{colon (:)}     & 61 & \multicolumn{1}{c}{colon (:)}     & 40  \\ 
                            & \multicolumn{1}{c}{semicolon (;) } & 61 & \multicolumn{1}{c}{semicolon (;) } & 39 \\ \hline
\end{tabular}}
\caption{BLEU Scores of translations for each delimiter}
\label{tab:bleu-table}
\end{table}

\noindent\textbf{Setup.} An important consideration for any de-biasing measure is to ensure the overall translation quality does not get impacted as an unwanted side effect. In our approach, we observed the choice of delimiter used during the addition of contexts has an impact on the translation quality. This was a critical deciding factor in choosing the right delimiter for our de-biasing approach, to ensure negligible impact on the translation quality while balancing for the precision in de-biasing.

For each of the language pairs, we draw a parallel corpus from Tatoeba\footnote{ \url{https://tatoeba.org/en/downloads}.} and draw 300 sentences from this dataset such that they contain the occupations used in the WinoMT dataset. We then translate these sentences after applying the top 50 contexts per sentence for each occupation and calculate the BLEU score after removing the contexts. We conduct our experiments testing the following delimiters: hash (\#), period (.), colon (:) and semicolon (;).

\noindent\textbf{Results.} We observe that when hash (\#) is used as a delimiter, the translation quality of the input sentence does not change compared to its translation without context (\autoref{tab:bleu-table}). However, the translation quality degrades when period, colon and semicolon are used as delimiters.
This is due to the fact that the delimiters period, and semicolon also naturally occur in the training corpora, leading to their presence within the output translation, making it hard for our post-processing pipeline to ascertain the exact boundary between the context and the sentence.

Thus, while we observe the best de-biasing performance by the use of the delimiter colon (:) (\autoref{tab:delimiters_table}), we recommend using hash (\#) as it provides competitive de-biasing performance while maintaining the best translation quality.

\begin{table}[ht]
\centering
\resizebox{0.48\textwidth}{!}{%
\begin{tabular}{cccc}

\hline
\multirow{2}{*}{Target Language} & \multirow{2}{*}{Delimiter} & OPUS-MT & M2M-100 \\ \cline{3-4} 
&       & $\mathcal{A}_{all}~(\%)$   & $\mathcal{A}_{all}~(\%)$       \\ \hline
\multirow{2}{*}{German(de)}      & colon (:)                      & \textbf{65.37}   & \textbf{62.51} 
                                 \\ 
                                 & semi-colon (;)                 & 64.00      & 62.04
                                 \\ 
                                 & hash (\#)                 & 62.09      & 58.25
                                 \\ 
                                 & period (.)                     & 61.39        & 59.73        \\ \hline
\multirow{2}{*}{French(fr)}      & colon (:)                      & 58.97   & 58.46   \\ 
                                 & semi-colon (;)                 & 58.26   & \textbf{58.53}
                                 \\ 
                                 & hash (\#)                 & \textbf{59.88}   & 54.03
                                 \\ 
                                 & period (.)                     & 56.49        &  57.25       \\ \hline
\multirow{2}{*}{Spanish(es)}     & colon (:)                      & 63.38   & \textbf{67.92}   \\ 
                                 & semi-colon (;)                 & \textbf{63.98}   & 66.73
                                 \\ 
                                 & hash (\#)                 & 58.97   & 59.25
                                 \\ 
                                 & period (.)                     & 60.45        & 59.40        \\ \hline
\end{tabular}
}

\caption{Results for different delimiters per model and per language pairs for the WinoMT dataset. $\mathcal{A}_{all}~(\%)$ reflects the average accuracy of correct gender associations for all templates applied to all sentences.}
\label{tab:delimiters_table}
\end{table}

\noindent\textbf{Takeaway.} \textit{The choice of delimiter governs the translation quality of the input sentence. Using hash(\#) as the delimiter provides the best trade-off between translation quality and removal of bias.}

\subsection{Are certain NMT models more sensitive towards contexts?}
\label{sec:css}
\begin{figure*}
     \centering
     \subfloat[OPUS-MT]{\includegraphics[trim={0, 0, 0, 80},clip, width=0.5\textwidth, height=3.8cm]{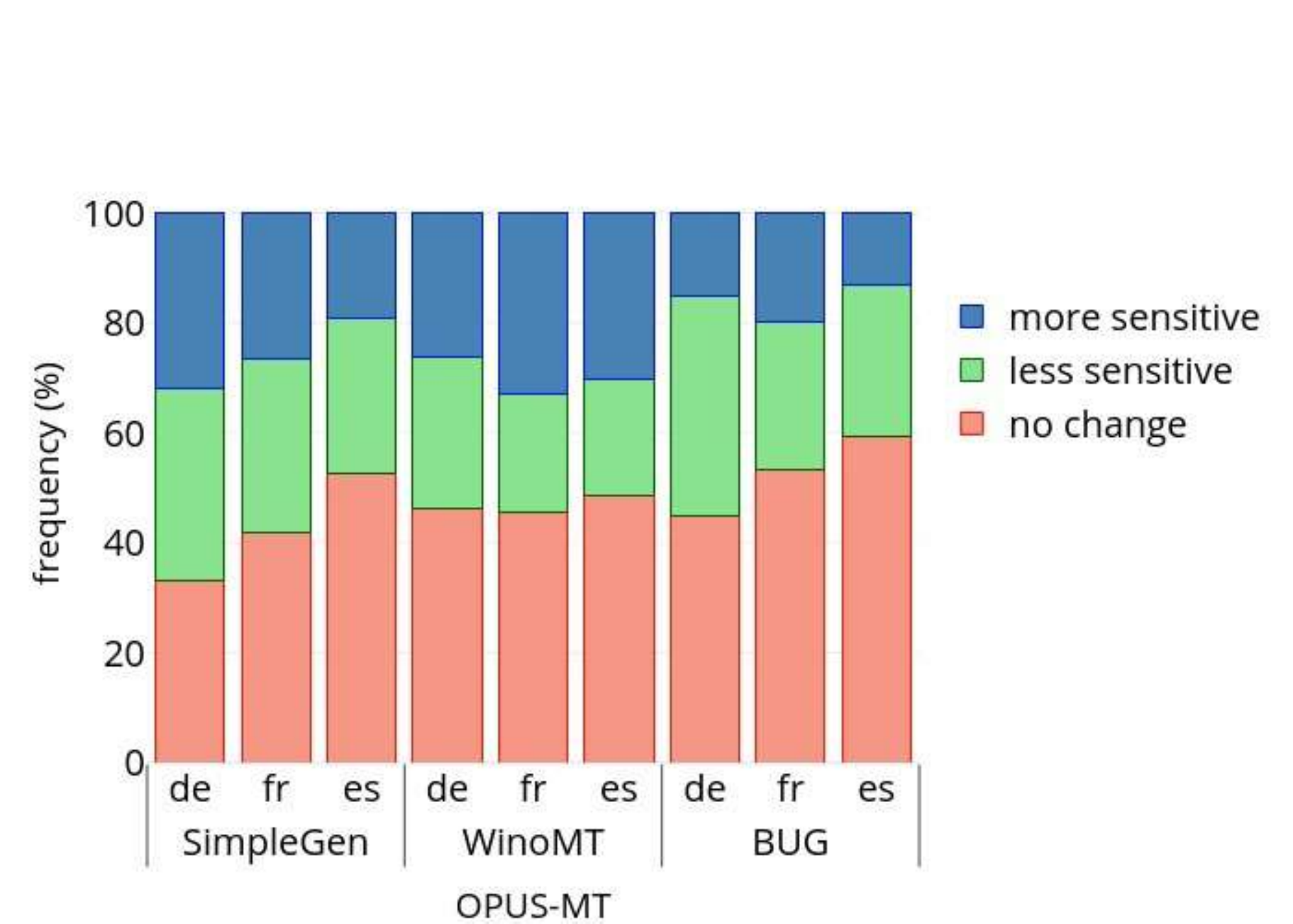}\label{opus_sensitivity}}
     \subfloat[M2M-100]{\includegraphics[trim={0, 0, 0, 80},clip, width=0.5\textwidth, height=3.8cm]{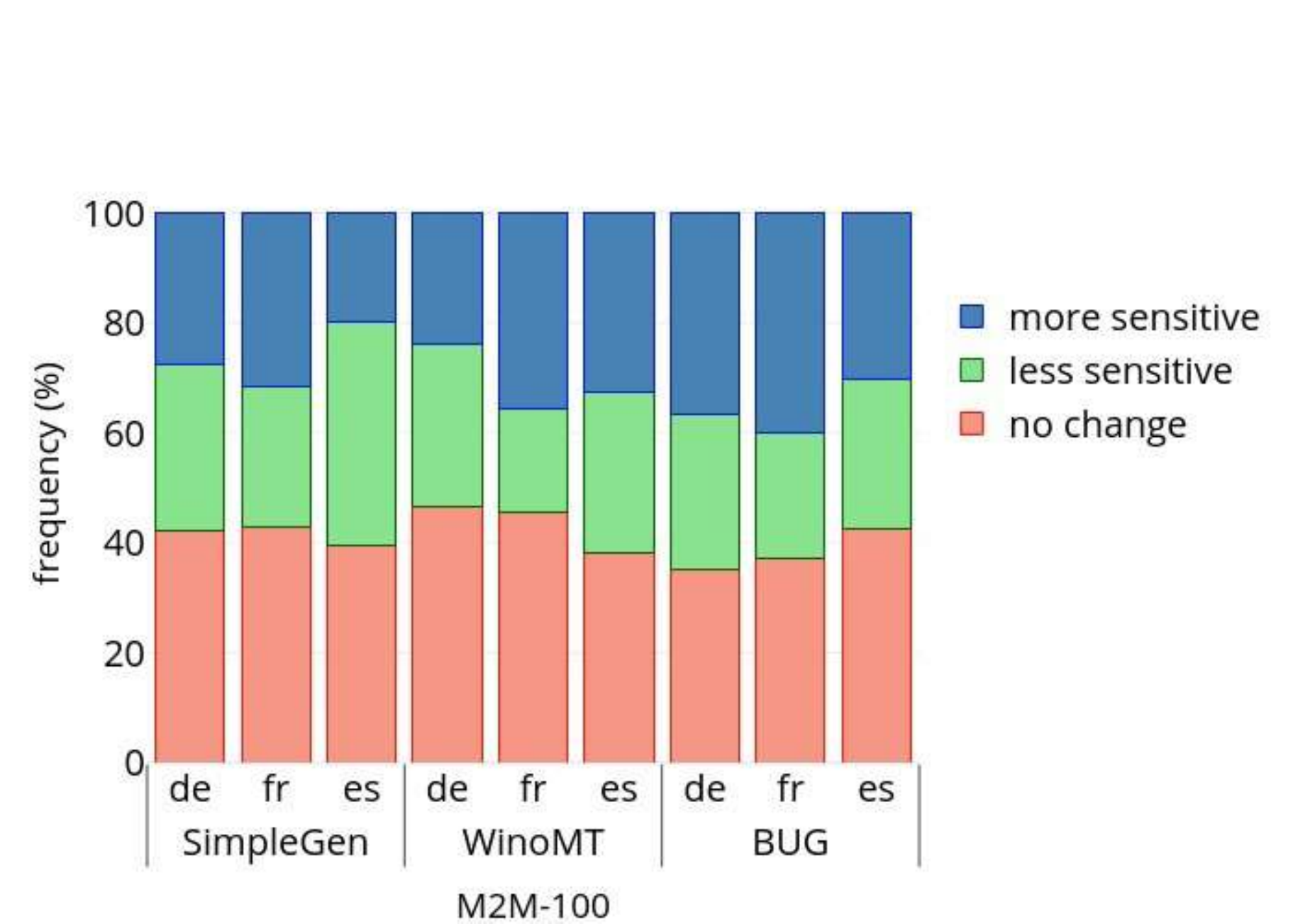}\label{m2m_sensitivity}}
     \caption{Distribution of sentences on the basis of sensitivity to the contexts for OPUS-MT and M2M-100 model. We define three bins: \textit{no change} (\texttt{CSS} score is 0), \textit{less sensitive}  (\texttt{CSS} Score <= 0.5) and \textit{more sensitive} (\texttt{CSS} Score > 0.5) on addition of contexts.}
     \label{fig:sensitivity}
\end{figure*}

\noindent\textbf{Setup.} In our preliminary experiments we observe that certain sentences are more sensitive to the addition of contexts than others. NMT systems reportedly are highly sensitive to modifications and perturbations in the input \cite{fadaee-monz-2020-unreasonable, dankers-etal-2022-paradox}. Thus, in this study, we aim to quantify the sensitivity of the translation model towards contexts. Concretely, we evaluate the sensitivity of a source sentence $X$ towards \textit{both} relevant and counterfactual contexts - i.e. applying $T$ unique templates with \textit{both} male and female gender signals, irrespective of the gender of the entity in the source sentence ($g_{X}$). Our objective is to observe if providing \textit{any} context triggers the model to \textit{change} the gender association of the entity in the translation, i.e., how \textit{sensitive} the model ($g_{\hat{Y}} \neq g_{\hat{Y}_{\not t}}, t \in T$) is. Thus, we define a context-sensitivity score (\texttt{CSS}) for sentences $X$ in dataset $D$ as the \textit{percentage of instances where the model changed the target entity gender association on the application of context}.

    \[\texttt{CSS}_{X} = \frac{1}{|D|} \sum_{X\in D}\frac{\sum_{t=1}^{T} I(g_{\hat{Y}} \neq g_{\hat{Y}_{\not t}})_{X,X_c \xrightarrow{\mathbb{M}} \hat{Y},\hat{Y}_{\not t}}}{2|T|}\]
where, $|T|$ is the total number of templates, used twice for male and female genders. We compute the \texttt{CSS} score for each language-model pair, and based on empirical evidence we categorize sentences into three distinct bins: \textit{no-change} for \texttt{CSS} score = 0, \textit{less sensitive}, for \texttt{CSS} $<= 0.5$; and \textit{more sensitive}, for \texttt{CSS} $> 0.5$.

\noindent\textbf{Results.} \autoref{fig:sensitivity} shows the frequency distribution of the sentences according to sensitivity range for each model and dataset. We observe M2M-100 model exhibiting higher sensitivity towards the addition of contexts. However, M2M-100 was also the most biased model before the application of contexts (\autoref{tab:full_table}). This result highlights that the many-to-many pre-training method (as used by M2M-100) is sensitive towards the change in input, which can be leveraged to develop a better, unbiased translation system using our method.

\noindent\textbf{Takeaway.} \textit{M2M-100 model proves to be more sensitive to contexts than OPUS-MT.}

\subsection{Does the NMT models understand the semantics of the context?}
\label{sec:semantics}

\begin{figure*}[h!]
     \centering
     \subfloat[OPUS-MT]{\includegraphics[width=0.5\textwidth]{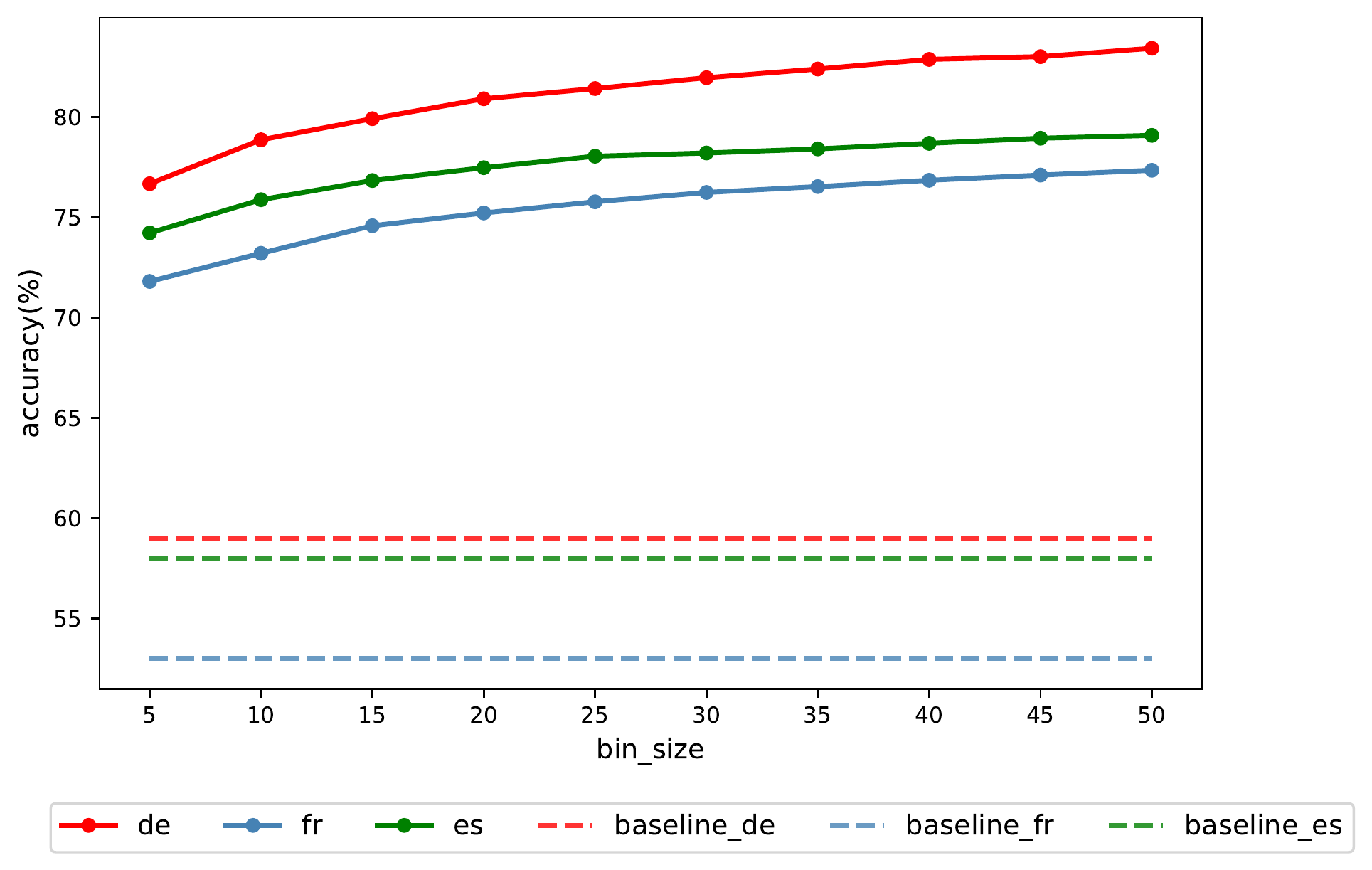}\label{opus_time_complexity}}
     \subfloat[M2M-100]{\includegraphics[width=0.5\textwidth]{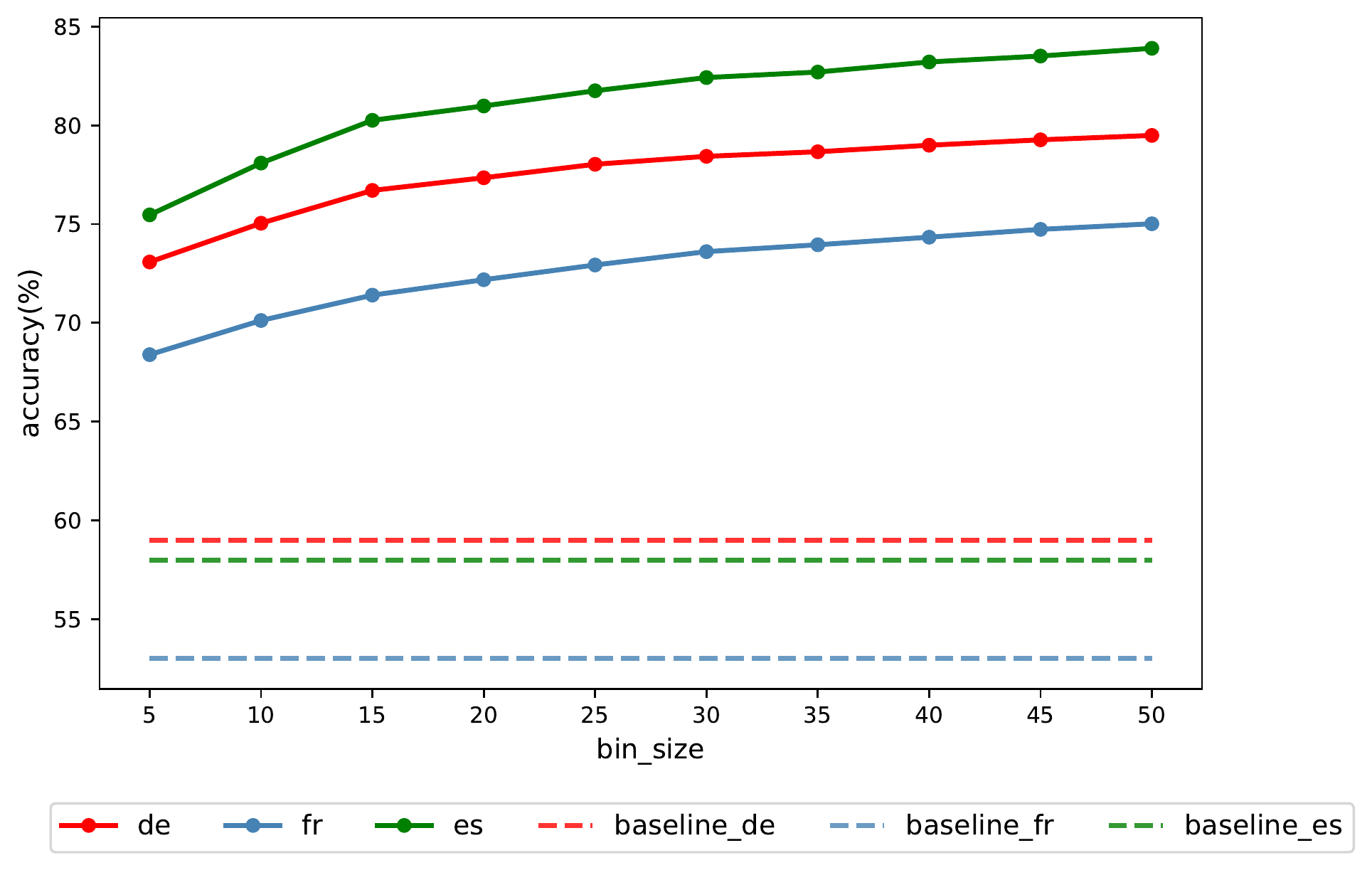}\label{m2m_time_complexity}}
     \caption{Variation in performance of the accuracy when using greedy strategy with increasing template space evaluated over WinoMT dataset. Baseline values represent the accuracy on the base dataset (with no context).}
     \label{fig:time_complexity}
\end{figure*}

\noindent\textbf{Setup.} While we observe marked improvements after adding contexts to the input, it is important to understand whether the semantics of the contexts matter for the model. We thus construct \textit{irrelevant} contexts containing no information about the gender of the target occupation, but having similar syntactic markup as $T$ \footnote{We provide the full list of irrelevant contexts in Appendix, \autoref{tab:irrelevant_contexts}}.
We then compute the average accuracy after the application of these contexts using the evaluation approach mentioned in \autoref{sec:average_acc}.

\noindent\textbf{Results.} We observe that adding gender-irrelevant contexts results do result in a significant decrease in accuracy (\autoref{tab:gi_table}). These results indicate that the contexts containing gender-relevant information are indeed useful for the model to de-bias the stereotypical occupation bias in translation, and irrelevant contexts hurt the performance by making the sentence more ambiguous for the NMT model.

\begin{table}[htb!]
\centering
\resizebox{0.48\textwidth}{!}{%
\begin{tabular}{ccccc} 
\hline
Target Language & Model & $\mathcal{A} (\%)$ & $\mathcal{A}_{all}~(\%)$ & $\mathcal{A}_{all\_gi}~(\%)$ \\ 
\hline
German(de) & \multirow{3}{*}{OPUS-MT} & 58.03 & 59.95 & 53.27 \\ 

French(fr) &  & 57.28 & 62.70 & 57.44 \\ 

Spanish(es) &  & 67.34 & 70.65 & 62.74 \\ 
\hline
German(de) & \multirow{3}{*}{M2M-100} & 54.05 & 56.10 & 53.09 \\ 

French(fr) &  & 53.34 & 60.00 & 55.23 \\ 

Spanish(es) &  & 59.75 & 64.22 & 58.05 \\
\hline
\end{tabular}
}
\caption{Results from adding gender irrelevant contexts to SimpleGen dataset. $\mathcal{A}$ represents the original accuracy of the dataset for the language model pairs. $\mathcal{A}_{all}$ is the average accuracy when correct contexts (with gender signals) are added, whereas  $\mathcal{A}_{all\_gi}$ represents the accuracy when gender-irrelevant contexts are added to the dataset.}
\label{tab:gi_table}
\end{table}

\noindent\textbf{Takeaway.} \textit{Both OPUS-MT and M2M-100 understand the meaning of the contexts.}

\subsection{Does the de-biasing accuracy vary with gender?}

\begin{table}[htb!]
\resizebox{\linewidth}{!}{%
\begin{tabular}{ccccc} 
\hline
Dimension & Gender & $\mathcal{A} (\%)$ & $\mathcal{A}_{all}~(\%)$ & $\delta (\%)$ \\ 
\hline
\multirow{2}{*}{Strong Stereotypes} & Female & 62.66 & 63.35 & 0.69 \\

 & Male & 56.76 & 59.98 & 3.22 \\ 
\hline
\multirow{2}{*}{Weak Stereotypes} & Female & 43.26 & 47.76 & 4.50 \\

 & Male & 77.81 & 76.45 & -1.36 \\
\hline
\end{tabular}
}
\caption{Comparision between male and female on gender on the basis of degree of stereotypes. $\mathcal{A}$ represents the aggregated accuracy of the original dataset without adding context. $\mathcal{A}_{all}$ is the aggregated accuracy with context added and $\delta$ represents the improvement in the accuracy after adding relevant contexts.}
\label{gender-tab}
\end{table}

\noindent\textbf{Setup.} In this section, we critically analyze the difference in the accuracy among different gender-stereotypical occupations before and after adding contexts. We hypothesize that the degree of stereotype of the occupation is an important factor to consider while applying a de-biasing technique, as certain occupations can be more challenging for a model to reduce its bias. For example, the sentence \textit{``That nurse is a funny man.''} is much harder to de-bias compared to the sentence \textit{``That teacher is a funny man.''}, as the societal stereotype of (\texttt{nurse}, \texttt{female}) is more prevalent in the data distribution.
%add one plot of \% females in a profession on x axis and accuracy before and after context on y axis (scatter plot aggregated across all languages and models in WinoMT dataset)
Thus, we perform the comparison among the degree of stereotypes in the occupations. We classify certain occupations which conform strongly to societal stereotypes using the WinoMT dataset \cite{stanovsky-etal-2019-evaluating} and the \href{https://www.census.gov/programs-surveys/cps.html}{US Current Population Survey (CPS)}. The remaining occupations are deemed as ``weakly stereotyped''. We compare the performance of our approach on these two bins.

\noindent\textbf{Results.} From our analysis (\autoref{gender-tab}), we first observe that before the addition of the context, the gender-association accuracy is better for strong female stereotypes than male stereotypes. However, post application of contexts, we observe larger improvement in male stereotypical occupations ($\delta = 3.22$) compared to their female counterparts ($\delta = 0.69$). The trend is opposite in case of weakly-stereotypical occupations. In fact, the gender association of male entities suffer post-addition of contexts, while female weakly-stereotypical entities are more correctly annotated ($\delta = 4.50$). Our results indicate that in the case of strong stereotypes, female occupations are much harder to de-bias, owing to their higher prevalence in the data used to train NMT models.

% \begin{itemize}
%     \item The original accuracy is better in case of female stereotypical professions as compared to male. However, the improvement($\delta$) is higher for male stereotypical occupations.
%     \item The sentences with male gold label perform better than their female counterparts, suggesting that models find it easier to translate sentences with male entities as opposed to female entities.
%     \item Interestingly, sentences with female entities had a greater improvement in accuracy after adding contexts when compared to their male counterparts.
% \end{itemize}
% The results of the two dimensions combined imply that models find it hard to translate male stereotypical profession when present in a female context. However, adding the right context aided their translations, thus leading to an improved accuracy.

\noindent\textbf{Takeaway.} \textit{Sentences having strongly stereotyped male-centric gender bias can be corrected more effectively by using contexts than their female counterparts.}

\subsection{What factors determine an effective template?}

\noindent\textbf{Setup.} We perform an empirical analysis to understand which factors determine the effectiveness of a template for correcting the gender bias in a sentence containing stereotypical occupation bias. To have a diverse set of contexts, we vary the templates with respect to the length of tokens, the number of gender signals (whether the gender is referred to by one or more nouns/pronouns) and the minimum distance of a gender signal from the target profession (i.e, how close the occupation word and the gender signal word(s) are in terms of token distance) \footnote{The full list of gender signal keywords and values used in the templates are described in Appendix \autoref{tab:temp_placeholder}.}.

% TODO rewrite this section
We compare the following properties of the templates: length of tokens ($l$), number of tokens with gender connotations ($s$) and the minimum distance from gender connotations from the token containing the target profession ($d$). For example, the context template \textit{``The }\{occupation\} \textit{in the next sentence identifies} \{m/f-ref-prn\} \textit{using the pronouns }\{m/f-sbj-prn\}/\{m/f-obj-prn\}'' has the features $s=2$, $d=3$ and $l=11$.

\noindent\textbf{Results.} As can be observed in \autoref{tab:pcc}, the number of gender signals ($s$) and token length ($l$) is positively correlated to the accuracy of the templates, i.e., longer templates with more number of gender signals lead to a larger improvement in the accuracy.
Interestingly, the relative distance(d) of the gender signals from the target profession token is negatively correlated with the accuracy, highlighting issues in co-reference resolution.

\noindent\textbf{Takeaway.} \textit{The token length and the number of gender signals used in the template is directly responsible for its effectiveness.}
 
%  We also analyzed if relative position of the context with respect to the original sentence affect the performance of the model. Since the decoding happens from left-to-right, we expected the contexts added before the sentence to perform better than the ones that appear after. While the average performance of the ones appearing before is slightly better than the templates that were added after, we do not see a significant difference in the performance of both.

\begin{table}
\centering
\resizebox{\linewidth}{!}{%
\begin{tabular}{ccccc} 
\hline
Model & Target Language & $d$ & $s$ & $l$ \\ 
\hline
\multirow{3}{*}{OPUS-MT} & German(de) & -0.18 & 0.27 & 0.13 \\ 

 & French (fr) & -0.16 & 0.26 & 0.15 \\ 

 & Spanish (es) & -0.20 & 0.23 & 0.06 \\ 
\hline
\multirow{3}{*}{M2M-100} & German(de) & -0.17 & 0.30 & 0.14 \\ 

 & French (fr) & -0.16 & 0.32 & 0.19 \\ 

 & Spanish (es) & -0.24 & 0.30 & 0.14 \\
\hline
\end{tabular}
}
\caption{Pearson correlation between the sentence accuracy of the templates for each language model pairs and various factors such as token length($l$) , number of gender signals($s$) and distance of the gender signal ($d$) from the profession across each language-model pair.}
\label{tab:pcc}
\end{table}

\subsection{What is the time complexity of the de-biasing pipeline?}
\label{sec:abl_time}

\noindent\textbf{Setup.} While our approach uses a simple methodology to de-bias translations, we understand that iterating through 50 contexts, as we do in our experiments, might seem relatively costlier in production systems.
To find the minimum size of the template set that can still reduce the bias in translations, we construct multiple subsets of randomly sampled templates, each having an increasing number of elements ranging from 5 to 50. 
We then perform our evaluation using the greedy strategy to choose a template from the given sample and evaluate it over the WinoMT dataset. We bootstrap the experiments 100 times to ensure statistical significance.

\noindent\textbf{Results.} While the performance does improve with an increase in sample size, even the smallest bin of 5 samples performs significantly well on the given dataset (\autoref{fig:time_complexity}). Thus, we highlight that while the performance of our approach is directly proportional to the number of templates considered, even smaller samples can also lead to considerable improvements in reducing the bias in translations.

\noindent\textbf{Takeaway.} \textit{Inference time complexity can be reduced by using less number of templates while maintaining competitive de-biasing accuracy.}

\section{Related Work}
% Several works have investigated bias in MT models and have demonstrated the asymmetries in the translations specific to gender. 

Several approaches to mitigate gender bias in Neural Machine Translation models have been proposed in the literature. \citet{escude-font-costa-jussa-2019-equalizing} de-bias pre-trained embeddings using hard de-biasing methods proposed by \citet{bolukbasi} which removes the gender associations from the representation of English gender-neutral words. However, the effectiveness of this approach has been debatable \citep{GONEN19,nissim2020fair,goldfarb-tarrant-etal-2021-intrinsic}. \citet{costa-jussa-de-jorge-2020-fine} propose a fine-tuning method using gender-balanced datasets containing an equal amount of masculine and feminine references and observe an improvement in the feminine forms.
% The approach, however, was not found to be as effective for gender translation on the anti-stereotypical WinoMT set. In our work, we perform the the evaluation over WinoMT dataset which consists of both pro-stereotypical as well as anti-stereotypical sentences and find improvements in both.
Closely related to our approach, \citet{vanmassenhove-etal-2018-getting} make use of gender tags (M or F) and prepend them to the source sentence during training and inference. \citet{stanovsky-etal-2019-evaluating} propose bias reduction with addition of pro-stereotypical adjectives. \citet{saunders-etal-2020-neural} explore the addition of gender tags at the word level. However, all of these approaches require the knowledge of gender metadata, which might not always be feasible to acquire. In our work, we bypass this limitation by using morphological taggers to extract the gender of the target entity. \citet{basta-etal-2020-towards} make use of the preceding sentence as context and concatenate it to the previous sentence. This context doesn't ensure gender-specific information and can be irrelevant with respect to the gender of the target entity. In our experiments, we find that such irrelevant contexts do not help with the de-biasing and in fact hurt the performance in some cases (\autoref{sec:semantics}). Contrary to this, we use relevant contexts which contain information about the gender of the target entity in the input. Our work is, to the best of our knowledge, the first attempt in exploring the effect of adding relevant sentences as contexts to de-bias translations during inference.
% We also study the sensitivity of translation models to such additional information.

\section{Conclusion}

In this work, we propose a simple and effective approach to correct stereotypical gender associations for occupations in translations during inference. Specifically, we add an unambiguous context to the input to state-of-the-art NMT models and observe that it enables the model to fix its own gender-bias towards gender-stereotyped occupations. Popular NMT models, such as M2M-100 and OPUS-MT, can effectively learn ``in-context'' how to correct their own biased translations provided the relevant context along the input.
% We also develop a mechanism to investigate the overall sensitivity of a sentence in a dataset per translation model and per target language towards a context.
Future work could consist of automatically choosing the correct template to add to the model during inference, or even generating ``prompts'' dynamically \cite{shin-etal-2020-autoprompt}.

\section*{Limitations}

\begin{itemize}
  \item \textit{Cost of iterating through template collection.} For larger models, iterating through a large set of templates to find the one that fixes the bias can be a costly process. Although we do analyze the complexity of our de-biasing pipeline in \autoref{sec:abl_time} and show that even a small set of templates can lead to significant improvement, iterating through a larger set for improved performance can be time-consuming.

  \item \textit{Accuracy of morph tagger.} An important limitation of our approach is that we rely on the accuracy of heuristic morphological taggers based on Stanza \cite{qi-etal-2020-stanza} to determine the gender of the source and target sentences before applying the corresponding context. Thus, a promising future work is to develop a robust gender extraction mechanism for effective de-biasing in complex sentence constructions.

  \item \textit{Proxy for Gender Bias.} Our approach evaluates gender bias using occupation, which are a commonly used proxy for gender in NLP (\citet{renduchintala-etal-2021-gender}, \citet{stanovsky-etal-2019-evaluating}, \citet{bolukbasi}, \citet{sharma2021evaluating}, \citet{Zhao2018GenderBI}). We acknowledge that this proxy might provide only a narrow view of gender bias in the NLP domain. Future work should investigate the removal of non-occupation gender-biases using our methodology.

  \item \textit{Applicability to only binary genders.} We use the existing gender-bias evaluation datasets that take only binary genders (man/woman) into consideration. Building upon this, our approach also takes the narrow view of binary gender. Furthermore, our reliance on US-Census-based occupations during evaluation might be covering only a limited set of occupations and stereotypes. We acknowledge these limitations and advocate that future work should consider non-binary gender as well as intersectional identities.

\end{itemize}

\section*{Ethics Statement}
Neural Machine Translation models have been shown to exhibit gender-bias, which also impacts their translation. Our work intends to investigate the usage of contextual information to fix this bias during inference. Our approach can thus contribute to the development of translation systems that are fairer and potentially less harmful. However, the removal of gender bias from translation is an active research problem, and our method can only improve bias to a certain extent. Special care should be taken while deploying NMT systems in production such that specific, harmful biases are pruned before they are served to the end user.

% Entries for the entire Anthology, followed by custom entries
\bibliography{anthology,custom_fixed}
\bibliographystyle{acl_natbib}

\clearpage
\appendix

\section{Appendix}
\label{sec:appendix}

\subsection{Evaluation dataset details}

Our evaluation was carried out on the following datasets.
\begin{enumerate}
    \item WinoMT - This dataset by (\citep{stanovsky-etal-2019-evaluating}) consists of 3888 sentences consisting of male, female and neutral entities. For our experiments, we filter out the neutral ones, leading to a total of 3648 sentences with 1826 male and 1822 female entities. 
    \item BUG - \citep{levy-etal-2021-collecting-large} (Balanced-Bug) is a large-scale corpus of 108K diverse real-world English sentences, collected via semi-automatic grammatical pattern matching to evaluate gender bias in various coreference resolution and machine translation models. Of these, we consider the sentences that have a complete overlap with the professions used in WinoMT dataset, leading to 3290 female and 3057 male entities.
    \item SimpleGEN - \citep{renduchintala-williams-2022-investigating} is a gender translation evaluation set based on gendered noun phrases in which there is a single, unambiguous, correct answer. There are a total of 2664 sentences with 1260 female and 1404 male entities.
\end{enumerate}

\subsection{Evaluating the performance of the heuristic morphological tagger}
As described in \autoref{sec:approach}, our greedy algorithm ($\mathcal{A}_c)$ heavily depends on the heuristic morphological tagger to extract the gender association($g_X$) from the source and from the translation ($g_{\hat{Y}}$). Our tagger makes use of the morphological tagger provided by Stanza \cite{qi-etal-2020-stanza} while using heuristics such as the presence of gender-specific words (e.g., he/him in English) to predict the gender.
To ensure that the gender predicted by this tagger is accurate, we measure the accuracy of the tagger in \autoref{tab:morph_tagger_acc}.
We observe the accuracy of the morphological tagger is specific to the dataset, and the performance of the same varies according to the complexity of sentence constructions and ambiguity within a dataset. Thus, our morphological tagger performs flawlessly on SimpleGen dataset, which is not surprising given the same dataset is constructed from artificially generated templates and thus exhibits the least ambiguity.

\begin{table}[ht]
\begin{center}
\begin{tabular}{ccccccc}
\hline
Dataset  & Accuracy (\%) \\ \hline
WinoMT     & 99.21   \\ 
BUG & 98.70  \\ 
SimpleGen & 100  \\ \hline
\end{tabular}
\end{center}
\caption{Accuracy of the custom tagger on various datasets for English (en)}
\label{tab:morph_tagger_acc}

\end{table}
In some cases, we observed that the heuristic tagger is unable to predict any gender, leading to the label \textit{unknown} (\autoref{tab:morph_tagger_acc}). Only a few professions (such as nurse) across particular languages appear to be affected by this issue. On average, we observe this issue in 5\% sentences in the entire datasets, with the highest in BUG dataset (7\%). Out of the two models, M2M-100 translations appear to display a higher propensity for this issue. In our experiments, these unknown labels contribute to the error of the model, and thus our method could be improved by the use of a better morphological tagger in the future.
\begin{table}[tb!]
\begin{center}
\begin{tabular}{cc}
\hline
Keywords &
  Values \\ \hline
\begin{tabular}[c]{@{}l@{}}f-n \\ m-n \\ f-n-pl \\ m-n-pl \\ f-sbj-prn \\ m-sbj-prn \\ f-n-sg \\ m-n-sg \\ f-pos-prn \\ m-pos-prn \\ f-obj-prn \\ m-obj-prn \\ f-ref-prn \\ m-ref-prn\end{tabular} &
  \begin{tabular}[c]{@{}l@{}}female \\ male \\ women \\ men \\ she \\ he \\ gal, woman \\ guy, man \\ her \\ his \\ her \\ him \\ herself \\ himself\end{tabular} \\ \hline
\end{tabular}
\end{center}
\caption{Key-Value pairs used to fill the placeholders in the templates while creating the contexts.}
\label{tab:temp_placeholder}
\end{table}

\begin{table*}[h]
\centering
\resizebox{\linewidth}{!}{%
\begin{tabular}{c|c|c|c|ccccc}
\hline
\multirow{2}{*}{Delimiter} &
  \multirow{2}{*}{Model} &
  \multirow{2}{*}{Dataset} &
  Without Context &
  \multicolumn{5}{c}{With Context} \\ \cline{4-9} 
   &
   &
   &
  \multicolumn{1}{c|}{$\mathcal{A} (\%)$} &
  \multicolumn{1}{c}{$\mathcal{A}_C (\%)$} &
  \multicolumn{1}{c}{$\mathcal{A}_all~(\%) $} &
  \multicolumn{1}{c}{$\texttt{CSS}$} &
  \multicolumn{1}{c}{$\mathcal{C}_U$} &
  $\mathcal{C}_L$ \\ \hline
\multirow{6}{*}{Period (.)} &
  \multirow{3}{*}{OPUS-MT} &
  WinoMT &
  \multicolumn{1}{c|}{59.45 (10.34)} &
  \multicolumn{1}{c}{\textbf{77.81}} &
  \multicolumn{1}{c}{59.44 (12.87)} &
  \multicolumn{1}{c}{0.20 (0.29)} &
  \multicolumn{1}{c}{59.00} &
  1.70 \\ 
 &
   &
  BUG &
  \multicolumn{1}{c|}{67.17 (19.33)}  &
  \multicolumn{1}{c}{\textbf{82.22}} &
  \multicolumn{1}{c}{65.16 (17.58)} &
  \multicolumn{1}{c}{0.21 (0.20)} &
  \multicolumn{1}{c}{55.39} &
  15.17 \\ 
 &
   &
  SimpleGen &
  \multicolumn{1}{c|}{60.88 (12.42)} &
  \multicolumn{1}{c}{\textbf{90.29}} &
  \multicolumn{1}{c}{66.18 (14.67)} &
  \multicolumn{1}{c}{0.24 (0.28)} &
  \multicolumn{1}{c}{75.94} &
  1.06 \\ \cline{2-8} 
 &
  \multirow{3}{*}{M2M-100} &
  WinoMT &
  \multicolumn{1}{c|}{54.9 (13.38)} &
  \multicolumn{1}{c}{\textbf{80.15}} &
  \multicolumn{1}{c}{58.67 (13.10)} &
  \multicolumn{1}{c}{0.21 (0.31)} &
  \multicolumn{1}{c}{57.84} &
  0.71 \\ 
 &
   &
  BUG &
  \multicolumn{1}{c|}{56.46 (14.98)} &
  \multicolumn{1}{c}{\textbf{82.54}} &
  \multicolumn{1}{c}{59.65 (16.92)} &
  \multicolumn{1}{c}{0.38 (0.41)} &
  \multicolumn{1}{c}{66.27} &
  23.70 \\ 
 &
   &
  SimpleGen &
  \multicolumn{1}{c|}{55.71 (7.76)} &
  \multicolumn{1}{c}{\textbf{94.82}} &
  \multicolumn{1}{c}{62.08 (11.00)} &
  \multicolumn{1}{c}{0.31 (0.69)} &
  \multicolumn{1}{c}{88.54} &
  1.51 \\ \hline
\multirow{6}{*}{Hash (\#)} &
  \multirow{3}{*}{OPUS-MT} &
  WinoMT &
  \multicolumn{1}{c|}{59.45 (10.34)} &
  \multicolumn{1}{c}{\textbf{77.52}} &
  \multicolumn{1}{c}{60.31 (10.70)} &
  \multicolumn{1}{c}{0.31 (0.41)} &
  \multicolumn{1}{c}{44.67} &
  8.13 \\ 
 &
   &
  BUG &
  \multicolumn{1}{c|}{67.17 (19.33)}  &
  \multicolumn{1}{c}{\textbf{82.75}} &
  \multicolumn{1}{c}{67.92 (17.10)} &
  \multicolumn{1}{c}{0.19 (0.31)} &
  \multicolumn{1}{c}{48.28} &
  8.87 \\ 
 &
   &
  SimpleGen &
  \multicolumn{1}{c|}{60.88 (12.42)} &
  \multicolumn{1}{c}{\textbf{84.38}} &
  \multicolumn{1}{c}{64.43 (13.12)} &
  \multicolumn{1}{c}{0.28 (0.38)} &
  \multicolumn{1}{c}{59.96} &
  10.63 \\ \cline{2-8} 
 &
  \multirow{3}{*}{M2M-100} &
  WinoMT &
  \multicolumn{1}{c|}{54.9 (13.38)} &
  \multicolumn{1}{c}{\textbf{80.94}} &
  \multicolumn{1}{c}{57.18 (9.99)} &
  \multicolumn{1}{c}{0.32 (0.40)} &
  \multicolumn{1}{c}{57.86} &
  14.05 \\ 
 &
   &
  BUG &
  \multicolumn{1}{c|}{56.46 (14.98)} &
  \multicolumn{1}{c}{\textbf{82.82}} &
  \multicolumn{1}{c}{65.78 (18.39)} &
  \multicolumn{1}{c}{0.37 (0.42)} &
  \multicolumn{1}{c}{63.75} &
  22.13 \\ 
 &
   &
  SimpleGen &
  \multicolumn{1}{c|}{55.71 (7.76)} &
  \multicolumn{1}{c}{\textbf{82.90}} &
  \multicolumn{1}{c}{60.11 (9.36)} &
  \multicolumn{1}{c}{0.28 (0.37)} &
  \multicolumn{1}{c}{61.71} &
  9.65 \\ \hline
\end{tabular}
}
\caption{Results aggregated over the language pairs (EN-DE, EN-FR,EN-ES) for each model/dataset. In ``Without Context", $\mathcal{A}$ reflects the accuracy of correct gender associations in the translation. In ``With Context", $\mathcal{A}_C$ is the accuracy after addition of extra-sentential context using the greedy strategy. $\mathcal{A}_{all}$ reflects the average accuracy of correct gender associations for all templates applied to all sentences, and $\texttt{CSS}$ represents the Context-Sensitivity score. The values in bracket are the standard deviation for the corresponding metric. $\mathcal{C}_U$ and $\mathcal{C}_L$ represent respectively the percentage of the biased sentences where at least one/all templates yields the correct prediction of the gender association.}
\label{tab:agg_table}
\end{table*}

\section{Computational Budget}
Our experiments are fairly lightweight in terms of the compute required, as we only run inference, and we avoid training any models. However, inference in NMT is a slow process due to beam  search (we used beam size 5), and thus translating  the three datasets along with all possible contexts requires approximately 1 day of GPU usage using  two NVIDIA P100 GPUs in parallel. We use the  HuggingFace \cite{huggingface} repository to run inference on the NMT models (OPUS-MT and M2M-100). 
% Code and data used will be released publicly after the review process.

% \input{tables/aggregate_table_new}

\begin{table*}[]
\begin{tabular}{c|c|c|cc|cc}
\hline
\multirow{2}{*}{Dataset}   & \multirow{2}{*}{Model}   & \multirow{2}{*}{Language} & \multicolumn{2}{c|}{Without Context}       & \multicolumn{2}{c}{With Context}          \\ \cline{4-7} 
                           &                          &                           & \multicolumn{1}{c}{F1\_Male} & F1\_Female & \multicolumn{1}{c}{F1\_Male} & F1\_Female \\ \hline
\multirow{6}{*}{WinoMT}    & \multirow{3}{*}{M2M-100} & German (de)               & \multicolumn{1}{c}{69.7}     & 39.2       & \multicolumn{1}{c}{67.2}     & 44.8       \\ 
                           &                          & French (fr)               & \multicolumn{1}{c}{65.1}     & 34.8       & \multicolumn{1}{c}{66.7}     & 28.3       \\ 
                           &                          & Spanish (es)              & \multicolumn{1}{c}{67.6}     & 43.2       & \multicolumn{1}{c}{67.3}     & 47.8       \\ \cline{2-7} 
                           & \multirow{3}{*}{OPUS-MT} & German (de)               & \multicolumn{1}{c}{66.7}     & 53.0       & \multicolumn{1}{c}{69.5}     & 51.4       \\ 
                           &                          & French (fr)               & \multicolumn{1}{c}{69.1}     & 34.9       & \multicolumn{1}{c}{70.6}     & 38.7       \\  
                           &                          & Spanish (es)              & \multicolumn{1}{c}{66.5}     & 51.1       & \multicolumn{1}{c}{66.1}     & 48.8       \\ \hline
\multirow{6}{*}{BUG}       & \multirow{3}{*}{M2M-100} & German (de)               & \multicolumn{1}{c}{62.7}     & 58.5       & \multicolumn{1}{c}{72.5}     & 64.5       \\ 
                           &                          & French (fr)               & \multicolumn{1}{c}{59.9}     & 49.0       & \multicolumn{1}{c}{65.5}     & 54.2       \\ 
                           &                          & Spanish (es)              & \multicolumn{1}{c}{65.6}     & 61.0       & \multicolumn{1}{c}{77.0}     & 69.7       \\ \cline{2-7} 
                           & \multirow{3}{*}{OPUS-MT} & German (de)               & \multicolumn{1}{c}{74.9}     & 69.7       & \multicolumn{1}{c}{70.6}     & 68.5       \\
                           &                          & French (fr)               & \multicolumn{1}{c}{62.8}     & 58.1       & \multicolumn{1}{c}{67.4}     & 59.5       \\  
                           &                          & Spanish (es)              & \multicolumn{1}{c}{78.8}     & 73.9       & \multicolumn{1}{c}{77.1}     & 76.2       \\ \hline
\multirow{6}{*}{SimpleGen} & \multirow{3}{*}{M2M-100} & German (de)               & \multicolumn{1}{c}{66.8}     & 31.4       & \multicolumn{1}{c}{63.8}     & 46.1       \\ 
                           &                          & French (fr)               & \multicolumn{1}{c}{67.4}     & 33.4       & \multicolumn{1}{c}{67.0}     & 49.8       \\
                           &                          & Spanish (es)              & \multicolumn{1}{c}{71.2}     & 39.8       & \multicolumn{1}{c}{71.9}     & 53.4       \\ \cline{2-7} 
                           & \multirow{3}{*}{OPUS-MT} & German (de)               & \multicolumn{1}{c}{67.3}     & 50.0       & \multicolumn{1}{c}{66.7}     & 58.0       \\ 
                           &                          & French (fr)               & \multicolumn{1}{c}{66.3}     & 48.0       & \multicolumn{1}{c}{66.5}     & 58.5       \\
                           &                          & Spanish (es)              & \multicolumn{1}{c}{72.3}     & 61.5       & \multicolumn{1}{c}{72.9}     & 68.5       \\ \hline
\end{tabular}
\caption{F1 scores aggregated over the language pairs (EN-DE, EN-FR,EN-ES) for each model/dataset. F1\_Male and F1\_Female represents the F1 scores of entities with gold label as male and female respectively. Scores before and after adding context are labelled as "Without Context" and "With Context" respectively}
\label{tab:f1}
\end{table*}

\begin{table*}
\centering
\resizebox{\textwidth}{!}{%
\begin{tabular}{l|l} 

\hline
A \{occupation\} appears in the following statement & In the next sentence, there is a revered \{occupation\} \\
The \{occupation\} in the next sentence is alive. & A well-known \{occupation\} is mentioned in the preceding statement. \\
The following statement refers to a respected \{occupation\} & A humble \{occupation\} is talked about in the following statement. \\
In the preceding sentence there is a \{occupation\} who is kind. & In the preceding sentence, there is a helpful \{occupation\} \\
The \{occupation\} in the next sentence is active. & The \{occupation\} in the next sentence works on the weekend. \\
The \{occupation\} in the preceding sentence is talented. & The \{occupation\} in the previous sentence is often praised by others. \\
The \{occupation\} in the following statement likes to travel. & An excellent \{occupation\} is mentioned in the following phrase. \\
There is a \{occupation\} in the next sentence who is trustworthy. & The \{occupation\} in the next sentence is short-tempered. \\
There is a loyal \{occupation\} in the next sentence. & In the previous sentence, there is a \{occupation\} who paints well. \\
The next sentence mentions a \{occupation\} & A \{occupation\} who is careful is mentioned in the previous phrase. \\
The \{occupation\} in the next sentence is energetic. & There is an accomplished \{occupation\} in the next sentence. \\
The person in the next sentence is a good \{occupation\} & The \{occupation\} in the next sentence will be on holiday next week. \\
The person in the previous sentence is a famous \{occupation\} & The following line is about an individual who works as a \{occupation\} \\
The next sentence mentions a \{occupation\} who is a learned professional. & The \{occupation\} in the following line is a nice person. \\
In the preceding sentence there is a funny \{occupation\} & A German \{occupation\} is mentioned in the previous line. \\
A renowned \{occupation\} is mentioned in the following statement. & The \{occupation\} in the previous sentence is fond of writing. \\
The \{occupation\} in the next sentence works hard. & A \{occupation\} is mentioned in the previous line who has two kids. \\
Everyone is fond of the \{occupation\} in the previous sentence. & The \{occupation\} in the next sentence likes gardening. \\
The \{occupation\} in the next sentence is afraid of dogs. & The \{occupation\} in the following statement is a great individual. \\
The next sentence talks about a \{occupation\} who is very polite. & The \{occupation\} in the next sentence knows Japanese. \\
There is a competent \{occupation\} in the next sentence. & There is a \{occupation\} in the previous statement who owns a car. \\
The \{occupation\} in the next statement is smart. & The \{occupation\} in the next sentence sings well. \\
The following sentence mentions a strict \{occupation\} & The \{occupation\} in the preceding sentence has diabetes. \\
In the next sentence, there is an experienced \{occupation\} & People talk highly about the \{occupation\} in the previous sentence. \\
The \{occupation\} in the next sentence is from India. & The \{occupation\} in the previous sentence was awarded. \\
\bottomrule
\end{tabular}
}
\caption{Templates used for creating gender irrelevant contexts}
\label{tab:irrelevant_contexts}
\end{table*}

\clearpage
\clearpage
\onecolumn
\begin{longtable}{p{0.7\textwidth}|p{0.75in}p{0.16in}p{0.16in}p{0.16in}}
\hline
\multicolumn{1}{c}{Templates used for creating contexts} &
  \multicolumn{1}{c}{$\mathcal{A}_{mean}$} &
  \multicolumn{1}{c}{$l$} &
  \multicolumn{1}{c}{$d$} &
  \multicolumn{1}{c}{$s$} \\ \hline
\endfirsthead
\multicolumn{5}{c}%
{{\bfseries Table \thetable\ continued from previous page}} \\
\hline
\multicolumn{1}{c}{Templates used for creating contexts} &
  \multicolumn{1}{c}{$\mathcal{A}_{mean}$} &
  \multicolumn{1}{c}{$l$} &
  \multicolumn{1}{c}{$d$} &
  \multicolumn{1}{c}{$s$} \\ \hline 
\endhead

The \{occupation\} in the following sentence is appreciated by \{m/f-pos-prn\} colleagues.                                & 68.82 (0.47) & 11 & 7  & 1 \\
While referring to \{m/f-ref-prn\}, the mentioned \{occupation\} uses the pronouns \{m/f-sbj-prn\}/\{m/f-obj-prn\}.       & 67.77 (0.47) & 11 & 3  & 1 \\
The \{occupation\} in the next sentence identifies \{m/f-ref-prn\} using the pronouns \{m/f-sbj-prn\}/\{m/f-obj-prn\}.    & 67.71 (0.43) & 12 & 5  & 2 \\
The \{occupation\} in the following sentence is excellent at \{m/f-pos-prn\} job.                                         & 65.95 (0.48) & 11 & 7  & 1 \\
The \{m/f-n\} \{occupation\} in the next sentence is a responsible \{m/f-n-sg\}.                                          & 65.85 (0.48) & 11 & 0  & 2 \\
There is a \{m/f-n\} \{occupation\} in the following sentence and \{m/f-sbj-prn\} is a polite \{m/f-n-sg\}.               & 65.8  (0.48) & 15 & 0  & 3 \\
The \{m/f-n\} \{occupation\} in the next sentence is valued at \{m/f-pos-prn\} workplace.                                 & 65.64 (0.48) & 12 & 0  & 2 \\
The \{m/f-n\} \{occupation\} in the next sentence is a \{m/f-n-sg\}.                                                      & 65.3  (0.48) & 10 & 0  & 2 \\
A talented \{occupation\} appears in the next sentence and \{m/f-sbj-prn\} identifies \{m/f-ref-prn\} as a \{m/f-n-sg\}.  & 65.21 (0.48) & 15 & 6  & 3 \\
The \{m/f-n\} \{occupation\} mentioned in the previous sentence is excellent at \{m/f-pos-prn\} job.                      & 64.96 (0.48) & 13 & 0  & 2 \\
A \{m/f-n\} \{occupation\} is being talked about in the next sentence.                                                    & 64.88 (0.48) & 11 & 0  & 1 \\
The \{occupation\} in the next sentence is liked by \{m/f-pos-prn\} coworkers.                                            & 64.78 (0.48) & 11 & 7  & 1 \\
In the following sentence there is a \{m/f-n\} \{occupation\} and \{m/f-sbj-prn\} is a humble \{m/f-n-sg\}.               & 64.75 (0.48) & 15 & 0  & 3 \\
In the following sentence is a \{m/f-n\} \{occupation\} and \{m/f-sbj-prn\} is a polite \{m/f-n-sg\}.                     & 64.54 (0.48) & 14 & 0  & 3 \\
The next sentence speaks of a \{m/f-n\} \{occupation\}.                                                                   & 64.49 (0.48) & 8  & 0  & 1 \\
The next sentence talks about a \{m/f-n\} \{occupation\}.                                                                 & 64.45 (0.48) & 8  & 0  & 1 \\
The \{occupation\} in the following sentence is the best among \{m/f-pos-prn\} peers.                                     & 64.4  (0.48) & 12 & 8  & 1 \\
The \{m/f-n\} \{occupation\} in the previous sentence is well-known for \{m/f-pos-prn\} expertise.                        & 64.32 (0.48) & 12 & 0  & 2 \\
The preceding sentence's \{m/f-n\} \{occupation\} is well-liked by \{m/f-pos-prn\} coworkers.                             & 64.14 (0.48) & 10 & 0  & 2 \\
The person in the following sentence is a \{m/f-n\} and is the only \{m/f-n\} \{occupation\} among \{m/f-pos-prn\} peers. & 64.13 (0.48) & 18 & 0  & 2 \\
A \{m/f-n\} \{occupation\} has been mentioned in the next sentence.                                                       & 64.09 (0.48) & 10 & 0  & 1 \\
The \{occupation\} in the following line enjoys \{m/f-pos-prn\} work.                                                     & 63.97 (0.48) & 9  & 5  & 1 \\
The \{occupation\} in the next sentence is fond of \{m/f-pos-prn\} job.                                                   & 63.45 (0.48) & 11 & 7  & 1 \\
The preceding sentence's \{m/f-n\} \{occupation\} is respected by \{m/f-pos-prn\} colleagues.                             & 63.39 (0.48) & 10 & 0  & 2 \\
The \{occupation\} in the next sentence loves \{m/f-pos-prn\} job.                                                        & 63.38 (0.48) & 9  & 5  & 1 \\
The following statement refers to a \{m/f-n\} \{occupation\} who is valued at \{m/f-pos-prn\} workplace.                  & 63.36 (0.49) & 14 & 0  & 2 \\
There is a \{occupation\} in the following sentence and \{m/f-pos-prn\} gender is \{m/f-n\}.                              & 63.36 (0.48) & 13 & 5  & 2 \\
A confident \{m/f-n\} \{occupation\} is being spoken about in the next sentence.                                          & 63.36 (0.48) & 12 & 0  & 1 \\
The \{m/f-n\} \{occupation\} who was mentioned in the preceeding phrase is well-known for \{m/f-pos-prn\} knowledge and experience. &
  63.31 (0.48) &
  17 &
  0 &
  2 \\
There is a \{m/f-n\} \{occupation\} in the next phrase, and \{m/f-sbj-prn\} is a nice individual.                         & 63.26 (0.49) & 15 & 0  & 2 \\
The \{occupation\} in the next sentence is talented and is excellent at \{m/f-pos-prn\} job.                              & 63.13 (0.47) & 14 & 10 & 1 \\
The \{occupation\} in the next sentence is great at \{m/f-pos-prn\} work.                                                 & 63.08 (0.48) & 11 & 7  & 1 \\
The \{occupation\} in the next sentence is great at \{m/f-pos-prn\} job.                                                  & 62.98 (0.48) & 11 & 7  & 1 \\
The person in the previous sentence is a \{m/f-n-sg\} and is the only \{m/f-n\} \{occupation\} in \{m/f-pos-prn\} group.  & 62.9 (0.49) & 18 & 0  & 3 \\
The preceding sentence talks about a \{m/f-n\} \{occupation\} who loves \{m/f-pos-prn\} job.                              & 62.73 (0.49) & 12 & 0  & 2 \\
The \{occupation\} in the next sentence is a smart \{m/f-n-sg\}.                                                          & 62.6 (0.48) & 10 & 7  & 1 \\
The individual in the preceding sentence is a \{m/f-n-sg\} and is the only \{m/f-n\} \{occupation\} amongst \{m/f-pos-prn\} peers. &
  62.47 (0.49) &
  18 &
  0 &
  3 \\
We are talking about a \{m/f-n\} \{occupation\} in the following sentence.                                                & 62.42 (0.49) & 11 & 0  & 1 \\
A \{m/f-n\} \{occupation\} who enjoys \{m/f-pos-prn\} work is described in the previous statement.                        & 62.37 (0.49) & 13 & 0  & 2 \\
The \{occupation\} in the previous sentence is respected by \{m/f-pos-prn\} colleagues.                                   & 62.33 (0.49) & 11 & 7  & 1 \\
A \{m/f-n\} \{occupation\} is shown in the previous statement, and \{m/f-sbj-prn\} is a proud \{m/f-n-sg\}.               & 62.23 (0.49) & 15 & 0  & 2 \\
The \{occupation\} in the previous sentence is known for \{m/f-pos-prn\} abilities and is a \{m/f-n-sg\}.                 & 62.18 (0.49) & 15 & 7  & 2 \\
In the preceding sentence is a \{m/f-n\} \{occupation\} who is valued at \{m/f-pos-prn\} workplace.                       & 62.11 (0.49) & 14 & 0  & 2 \\
The \{occupation\} in the previous sentence is known for \{m/f-pos-prn\} skills and \{m/f-sbj-prn\} is a \{m/f-n-sg\}.    & 62.02 (0.49) & 16 & 7  & 3 \\
A \{m/f-n\} \{occupation\} appears in the following statement.                                                            & 61.99 (0.49) & 8  & 0  & 1 \\
The next sentence talks about a \{m/f-n-sg\} and \{m/f-sbj-prn\} is a \{occupation\}.                                     & 61.8  (0.49) & 12 & 2  & 2 \\
The following sentence is about a \{m/f-n-sg\} whose occupation is \{occupation\}.                                        & 61.71 (0.49) & 11 & 3  & 1 \\
According to the previous sentence, the \{m/f-n-sg\} is a \{m/f-n\} \{occupation\}.                                       & 61.68 (0.49) & 11 & 0  & 2 \\
In the previous statement, a \{occupation\} is mentioned, and \{m/f-sbj-prn\} is a \{m/f-n-sg\}.                          & 61.63 (0.49) & 13 & 3  & 2 \\
In the following phrase is a modest \{m/f-n\} \{occupation\}.                                                             & 61.55 (0.49) & 9  & 0  & 1 \\
In the previous line, the \{occupation\} is a \{m/f-n\} who refers to \{m/f-ref-prn\} as \{m/f-sbj-prn\}/\{m/f-obj-prn\}. & 61.15 (0.49) & 15 & 2  & 3 \\
In the next sentence is a \{m/f-n-sg\} and \{m/f-sbj-prn\} has been employed as a \{occupation\}.                         & 61.01 (0.49) & 15 & 5  & 2 \\
A \{m/f-n\} \{occupation\} was mentioned in the previous statement.                                                       & 60.95 (0.49) & 9  & 0  & 1 \\
In the previous sentence, the \{occupation\} identifies \{m/f-ref-prn\} as a \{m/f-n-sg\}.                                & 60.95 (0.49) & 11 & 1  & 2 \\
The next line is about a \{m/f-n-sg\} who works as a \{occupation\}.                                                      & 60.94 (0.49) & 12 & 4  & 1 \\
The \{occupation\} in the previous line is a \{m/f-n-sg\} of confidence.                                                  & 60.87 (0.49) & 11 & 6  & 1 \\
The \{occupation\} in the next sentence uses the pronouns \{m/f-sbj-prn\}/\{m/f-obj-prn\}.                                & 60.83 (0.49) & 10 & 7  & 1 \\
A \{m/f-n\} \{occupation\} appears in the previous statement.                                                             & 60.77 (0.49) & 8  & 0  & 1 \\
The previous statement \{m/f-n-pl\}tions a \{occupation\} and \{m/f-sbj-prn\} is a \{m/f-n-sg\}.                          & 60.72 (0.49) & 11 & 1  & 2 \\
The \{occupation\} in the preceding line is a trustworthy \{m/f-n-sg\}.                                                   & 60.7  (0.49) & 10 & 7  & 1 \\
The mentioned \{occupation\} and \{m/f-pos-prn\} colleagues are honest \{m/f-n-pl\}.                                      & 60.69 (0.49) & 9  & 1  & 2 \\
In the previous sentence the \{occupation\} identifies \{m/f-ref-prn\} as a \{m/f-n-sg\}.                                 & 60.62 (0.49) & 11 & 1  & 2 \\
In the previous sentence, there is a \{m/f-n\} \{occupation\}.                                                            & 60.59 (0.49) & 9  & 0  & 1 \\
The \{m/f-n-sg\} in the previous sentence is a responsible \{occupation\}.                                                & 60.43 (0.49) & 10 & 7  & 1 \\
The coworkers of the mentioned \{m/f-n\} \{occupation\} are also \{m/f-n-pl\}.                                            & 60.39 (0.49) & 10 & 0  & 2 \\
The previous sentence is about a \{m/f-n\} \{occupation\}.                                                                & 60.31 (0.49) & 8  & 0  & 1 \\
As per the previous sentence, there is a \{m/f-n\} \{occupation\}.                                                        & 60.27 (0.49) & 10 & 0  & 1 \\
The \{occupation\} in the next sentence is a \{m/f-n-sg\} and \{m/f-sbj-prn\} uses the pronouns \{m/f-sbj-prn\}/\{m/f-obj-prn\}. &
  60 (0.49) &
  15 &
  6 &
  3 \\
The \{occupation\} mentioned here is a polite \{m/f-n-sg\}.                                                               & 59.92 (0.49) & 8  & 5  & 1 \\
The following line is about a \{m/f-n-sg\} who works as a \{occupation\}.                                                 & 59.77 (0.49) & 12 & 4  & 1 \\
The following statement refers to a \{m/f-n-sg\} who works as a \{occupation\}.                                           & 59.74 (0.49) & 12 & 4  & 1 \\
The \{m/f-n-sg\} in the previous sentence is a sincere \{occupation\}.                                                    & 59.61 (0.49) & 10 & 7  & 1 \\
The \{m/f-n-sg\} in the previous sentence is a loyal \{occupation\}.                                                     & 59.58 (0.49) & 10 & 7  & 1 \\
There is a \{occupation\} in the next sentence and \{m/f-pos-prn\} gender is \{m/f-n\}.                                   & 59.58 (0.49) & 13 & 5  & 2 \\
The preceding sentence describes a \{m/f-n-sg\} \{occupation\} who likes \{m/f-pos-prn\} profession.                      & 59.58 (0.49) & 11 & 0  & 2 \\
The \{occupation\} in the previous sentence identifies \{m/f-ref-prn\} as a \{m/f-n-sg\}.                                 & 59.15 (0.49) & 11 & 5  & 2 \\
In the next sentence there is a \{occupation\} and \{m/f-pos-prn\} gender is \{m/f-n\}.                                   & 59.09 (0.49) & 13 & 1  & 2 \\
The \{occupation\} in the previous sentence is a \{m/f-n\} \{occupation\}.                                                & 59.02 (0.49) & 10 & 0  & 1 \\
The \{occupation\} in the previous sentence is respected at \{m/f-pos-prn\} workplace.                                    & 58.49 (0.5)  & 11 & 7  & 1 \\
Here, the \{occupation\} is a \{m/f-n-sg\}.                                                                               & 57.9  (0.5)  & 6  & 2  & 1 \\
The \{occupation\} in the previous sentence is a \{m/f-n-sg\} and \{m/f-sbj-prn\} likes \{m/f-pos-prn\} job.              & 57.8  (0.5)  & 14 & 6  & 3 \\
The \{occupation\} here is a confident \{m/f-n-sg\}.                                                                      & 56.93 (0.5)  & 7  & 4  & 1 \\
The \{occupation\} mentioned here is a humble \{m/f-n-sg\}.                                                               & 56.71 (0.49) & 8  & 5  & 1 \\
The correct gender of the \{occupation\} in the next sentence is \{m/f-n\}.                                               & 56.33 (0.5)  & 12 & 5  & 1 \\
Here, the \{occupation\} is a \{m/f-n-sg\} and uses the pronouns \{m/f-sbj-prn\}/\{m/f-obj-prn\}.                         & 53.94 (0.49) & 11 & 2  & 2 \\
The gender of the \{occupation\} in the next sentence is \{m/f-n\}.                                                       & 51.63 (0.49) & 11 & 5  & 1 \\
The pronouns \{m/f-sbj-prn\} and \{m/f-obj-prn\} are used by the \{occupation\} in the following phrase.                  & 50.34 (0.5)  & 14 & 4  & 2
\\ \hline
\caption{List of templates used to create the contexts for our evaluation.$\mathcal{A}_{mean}$ represents the accuracy for each template (across all lang-model pairs). The values in brackets represent the corresponding standard deviation. ($l$), ($s$), ($d$) represent token length, number of signals and minimum distance of a gender signal from the target profession respectively.The templates are sorted in decreasing order of the mean sentence accuracy.}
\label{tab:contexts-table}\\
\end{longtable}
\clearpage
\twocolumn

\end{document}